\pdfoutput=1

\documentclass[runningheads]{llncs}

\usepackage{color}

\usepackage{bm}

\usepackage{xspace,amssymb,amsmath,graphicx,srcltx,dblfloatfix,url}

\usepackage{amsmath,amscd,amsthm}
\usepackage{graphicx}
\graphicspath{{./}{./Fig/}{./Figs/}}

\usepackage[algo2e,vlined,algoruled]{algorithm2e}
\usepackage{algorithm}
\begin{document}

%

%
\pagestyle{headings}
\mainmatter


\renewcommand{\u}{\mathbf u}
\renewcommand{\v}{\mathbf v}

\renewcommand{\a}{\mathbf a}

\newcommand{\x}{\mathbf x}
\newcommand{\w}{\mathbf w}

\newcommand{\q}{\mathbf q}
\newcommand{\z}{\mathbf z}

\newcommand{\e}{\boldsymbol e}

\newcommand{\A}{\mathbf A}
\newcommand{\C}{\mathbf C}
\newcommand{\X}{\mathbf X}
\renewcommand{\H}{\mathbf H}

\newcommand{\eg}{{\em e.g.}}

\newcommand{\brho}{\boldsymbol \rho}

\newcommand{\Q}{\mathbf Q}

\def\psd{\succcurlyeq}  
\def\nsd{\preccurlyeq} 
\def\pd{ \succ } 
\def\nd{ \prec } 

\newcommand{\cH}{\mathcal H}
\newcommand{\cX}{\mathcal X}

\def\T{{\!\top}}
\def\Real{\mathbb{R}}
 
\newcommand{\norm}[2][2]{\ensuremath{ \left\| #2 \right\|_{ \mathrm{#1} } } }

\newcommand{\ADot}{ \ensuremath{  - \,} }

\def\NICTAFunding{{NICTA is funded by the Australian Government as represented by the Department of
Broadband, Communications and the Digital Economy and the Australian Research Council through the
ICT Centre of Excellence program.}}

%
%
\newcommand{\comment}[1]{}

\theoremstyle{plain}

%
%
\newcommand{\bSigma}{{\bf \Sigma}}
\def\Phi{{\phi }}

\title{LACBoost and FisherBoost: 
       Optimally Building Cascade Classifiers} 

%
%

\titlerunning{LACBoost and FisherBoost: 
Optimally Building Cascade Classifiers}

\authorrunning{C. Shen, P. Wang, and H. Li}
 \author{Chunhua Shen\inst{1,2} \and Peng Wang\inst{3}\thanks{P. Wang's
 contribution was made when visiting NICTA and Australian National University.}
 \and Hanxi Li\inst{2,1}}
 \institute{NICTA\thanks{\NICTAFunding}, Canberra Research Laboratory, ACT 2601, Australia
 \and Australian National University,  ACT 0200, Australia
 \and Beihang University, Beijing 100191, China}

\maketitle

\begin{abstract}

        Object detection is one of the key tasks in computer vision. 
        The cascade framework of Viola and Jones has become the 
        {\em de facto} standard. A classifier in each node of the cascade
        is required to achieve extremely high detection rates, instead of
        low overall classification error. 
        Although there are a few reported methods addressing  
        this requirement in the context of object detection,
        there is no a principled feature selection method that
        explicitly takes into account this asymmetric node learning objective. 
        We provide such a boosting algorithm in this work.
        It is inspired by the linear asymmetric classifier (LAC) of 
        \cite{Wu2005Linear} in that our boosting algorithm optimizes 
        a similar cost function. The new totally-corrective boosting  
        algorithm is implemented by the 
        column generation technique in convex optimization. 
        Experimental results on face detection suggest that
        our proposed boosting algorithms can improve the state-of-the-art 
        methods in detection performance.

\end{abstract}

\section{Introduction}

        Real-time detection of various categories of objects in images is one of the
        key tasks in computer vision.
        This topic has been extensively studied in the past a few years due to its
        important applications in surveillance, intelligent video analysis {\em etc}.  
        %
        %
        Viola and Jones proffered the first real-time face detector
        \cite{Viola2004Robust,Viola2002Fast}.
        To date, it is still considered one of the state-of-the-art, and their framework 
        is the basis of many incremental work afterwards.  
        Object detection is a highly asymmetric classification
        problem with the exhaustive scanning-window search being used to locate the target in an
        image. Only a few are true target objects among the millions of scanned patches.
        Cascade classifiers 
       %
       %
        have been proposed for efficient detection, which takes the asymmetric
        structure into consideration. 
        Under the assumption of each node of the cascade classifier makes independent classification
        errors, the detection rate and false positive rate of the entire cascade are:
        $ F_{\rm dr} = \prod_{ t =1}^N d_t    $  and
        $ F_{\rm fp} = \prod_{ t =1}^N f_t    $, respectively.
        As pointed out in \cite{Viola2004Robust,Wu2005Linear}, these two equations 
        suggest a {\em node learning objective}: 
        Each node should have an extremely high detection rate $d_t $ 
        ({\em e.g.}, $99.7\%$) and 
        a moderate false positive rate $ f_t $ ({\em e.g.}, $50\%$). 
        With the above values of $  d_t $ and $ f_t $, assume that 
        the cascade has $ N = 20 $ nodes, then $ F_{\rm dr} \approx  94\%$
        and $ F_{\rm fp} \approx  10^ {-6} $, which is usually the design goal.

        A drawback of standard boosting like AdaBoost is that it does not 
        take advantage of
        the cascade classifier.  AdaBoost only minimizes the overall classification error and does
        not minimize the number of false negatives.  
        In this sense, the features selected are not optimal for 
        the purpose of rejecting negative examples.
        At the feature selection and classifier training level, Viola and Jones
        leveraged the asymmetry property, to some extend, by
        replacing AdaBoost with AsymBoost \cite{Viola2002Fast}.
        AsymBoost incurs more loss for misclassifying a positive example by simply
        modifying AdaBoost's exponential loss. 
        Better detection rates were observed over the standard AdaBoost. Nevertheless, 
        AsymBoost addresses the node learning goal {\em indirectly}
        and still may not be the optimal solution.
        Wu {\em et al.} explicitly studied the node learning goal and they proposed 
        to use linear asymmetric classifier (LAC) and Fisher linear discriminant analysis (LDA)
        to adjust the linear coefficients of the selected weak classifiers
        \cite{Wu2005Linear,Wu2008Fast}.
        Their experiments indicated that with this post-processing technique, 
        the node learning objective can be better met, which is translated into improved 
        detection rates. 
        In Viola and Jones' framework, boosting is used to select 
        features and at the same time to
        train a strong classifier. Wu {\em et al.}'s work separates these two tasks:
        they still use AdaBoost or AsymBoost to select features; and at the second step, 
        they build a strong classifier using LAC or LDA. 
        Since there are two steps here, in Wu {\em et al.}'s work 
        \cite{Wu2005Linear,Wu2008Fast}, 
        the node learning objective is only considered at the 
        second step. At the first step---feature selection---the node learning objective is
        not explicitly considered. We conjecture  that
        {\em further improvement may be gained
        if the node learning objective is explicitly 
        taken into account at both steps}. 
        We design new  boosting algorithms to implement this idea and verify
        this conjecture. 
        Our major contributions are as follows.  
        \begin{enumerate}
           \item
                We develop new boosting-like algorithms via directly
                minimizing the objective function of 
                linear asymmetric classifier, which is termed as LACBoost (and
                FisherBoost from Fisher LDA). Both of them can be used to
                select features that is optimal for achieving the node learning goal in training a 
                cascade classifier.  To our knowledge, this is the first attempt to design such a 
                feature selection method. 
           \item
                LACBoost and FisherBoost share similarities with LPBoost 
                \cite{Demiriz2002LPBoost} in the sense that both use
                column generation---a technique originally proposed
                for large-scale linear programming (LP). 
                Typically, the Lagrange dual problem 
                is solved at each iteration in column generation. We instead solve
                the primal quadratic programming (QP) problem, which has a special structure
                and entropic gradient (EG)
                can be used to solve the problem very efficiently. 
                Compared with general interior-point based QP solvers, EG is much faster. 
                Considering one needs to solve QP problems a few thousand times for training
                a complete cascade detector, the efficiency improvement is enormous. 
                Compared with training an AdaBoost based cascade detector, the time needed 
                for LACBoost (or FisherBoost) is comparable.
                This is because for both cases, the majority of the time is spent on weak
                classifier training and bootstrapping. 
           \item
                 We apply LACBoost and FisherBoost to face detection and better performances are observed over
                 the state-of-the-art methods \cite{Wu2005Linear,Wu2008Fast}. The results
                 confirm our conjecture and show the effectiveness of LACBoost and FisherBoost. 
                 LACBoost can be immediately applied to other asymmetric classification problems.
           \item
                We also analyze the condition that makes the validity of LAC,
                and show that the multi-exit cascade might be more suitable
                for applying LAC learning of \cite{Wu2005Linear,Wu2008Fast} (and our LACBoost)
                rather than Viola-Jones standard cascade.
         \end{enumerate}
         Besides these, the LACBoost/FisherBoost algorithm differs from traditional boosting algorithms
         in that LACBoost/FisherBoost does not minimize a loss function. This opens new possibilities
         for designing new boosting algorithms for special purposes. 
         We have also extended column generation for optimizing nonlinear 
         optimization problems.            
         Next we review some related work that is closest to ours.

      %

        \textbf{Related work}
        There is a large body of  previous work in object detection
        \cite{pham07,Wu2003Rare};
        of particular relevance to our work is boosting object detection originated
        from Viola and Jones'
        framework.
        There are three important components that make Viola and Jones' framework
        tremendously successful:
        (1) The cascade classifier that efficiently filters out most negative patches
            in early nodes; and also contributes to enable the final classifier to have
            a very high detection rate;
        (2) AdaBoost that selects informative features and at the same time trains 
            a strong classifier;
        (3) The use of integral images, which makes the computation of Haar features 
            extremely fast. 
        Most of the work later improves one or more of these three components. 
        In terms of the cascade classifier, a few different approaches such as 
        soft cascade \cite{Bourdev05SoftCascade}, dynamic cascade 
        \cite{Rong2007}, and multi-exit cascade \cite{pham08multi}. 
        We have used the multi-exit cascade in this work.
        The multi-exit  cascade tries to improve the classification performance by 
        using all the selected weak classifiers for each node. 
       %
       %
        So for the  $ n $-th strong classifier (node), it uses all the weak classifiers
        in this node as well as those in the previous $ n - 1 $ nodes. 
        We show that the LAC post-processing can enhance the multi-exit cascade.
        More importantly, we show that the multi-exit cascade better meets LAC's 
        requirement of data being Gaussian distributions. 
        
        The second research topic is the learning algorithm 
        for constructing a classifier. 
        Wu {\em et al.} use fast forward feature selection to 
        accelerate the training procedure
        \cite{Wu2003Rare}. They have also proposed LAC to learn a better strong
        classifier \cite{Wu2005Linear}. 
        Pham and Cham recently proposed online asymmetric boosting 
        with considerable improvement in training time \cite{pham07}.
        By exploiting the feature
        statistics, they have also designed a fast method to train weak classifiers
        \cite{Pham2007Fast}. 
        Li {\em et al.} advocated FloatBoost to discard
        some redundant weak classifiers during AdaBoost's    
        greedy selection procedure \cite{Li2004Float}. 
        Liu and Shum proposed KLBoost to select features and train a strong 
        classifier \cite{Liu2003KL}. 
        Other variants of boosting have been applied to detection. 
        \comment{
        For example,
        Promising results were reported with
        LogitBoost \cite{Tuzel2008PAMI} that employs the logistic regression 
        loss, and GentleBoost \cite{Torralba2007} that uses adaptive 
        Newton steps to fit the additive model.

        New features have also been designed for improving the detection
        performance. Viola and Jones' 
        Haar features are not sufficiently discriminative for detecting
        more complex objects like pedestrians, or multi-view faces. 
        Covariance features \cite{Tuzel2008PAMI} and histogram of oriented gradients 
        \cite{Dalal2005HOG} have been proposed in this context. Both of them are
        possible to the use the idea of integral images/histograms to reduce the
        computation complexity. 
        }

        \textbf{Notation} 
        The following notation is used. 
        A matrix is denoted by a bold upper-case
        letter ($\X$); a column vector is denoted by a bold lower-case
        letter ($ \x $). 
        The $ i$th row of $\X $ is denoted by $ \X_{i:} $ 
        and the $ i $-th column $ \X_{:i}$.
        The identity matrix is $ \bf I $ and its size should be clear
        from the context. $ \bf 1 $ and  
        $ \bf 0 $ are column vectors of $ 1$'s and $ 0$'s,
        respectively.  
        We use $ \psd, \nsd $ to denote component-wise inequalities.

        Let $ \{ (\x_i, y_i  ) \}_{i = 1, \cdots, m}$ be the set of
        training data, where $ \x_i \in \cX$ and $ y_i \in \{-1,+1\}
        $, $ \forall i$. 
        The training set consists of $ m_1 $ positive training points
        and $ m_2 $ negative ones; $ m_1 + m_2 = m $. 
        Let $ h ( \cdot  ) \in \cH $ be a weak
        classifier that projects an input vector $ \x $ into 
        $\{-1, +1 \}$. Here we only consider discrete classifier
        outputs.
        We assume that the set $ \cH $ is finite and we
        have $ n $ possible weak classifiers. Let the matrix $ \H \in
        \Real^{ m \times n }$ where the $ (i,j)$ entry of $ \H $ is
        $ \H_{ij} =  h_j ( \x_i ) $. 
        $ \H_{ij} $ is the label predicted by weak classifier $
        h_j(\cdot) $ on the training datum $ \x_i $. 
        \comment{
        Therefore each
        column $ \H_{ :j }  $ of the matrix $ \H $ consists of the
        output of weak classifier $ h_j(\cdot) $ on all the training
        data; while each row $ \H_{ i: } $ contains  the outputs of all
        weak classifiers on the training datum $ \x_i $. 
        }
        We define a matrix $ \A \in  \Real^{ m \times n }$ such that its
        $( i, j )$ entry is
        $ \A_{ij} = y_i h_j (  \x_i ) $.
        \comment{
        Boosting algorithms entirely depend on the matrix $  \A  $ and
        do not directly interact with the training examples. 
        Our following discussion will largely focus on the matrix $\A$.
        We write the vector obtained by multiplying a matrix $  \A  $ 
        with a vector 
        $ \w $
        as $ \A \w $ and its $i$-th entry as $ (\A \w)_i$, which is the margin
        of the training datum $ \x_i $: $ \rho_i = \A_{ i :} \w = (\A \w)_i $.
        }

        \comment{
        The paper is organized as follows. We briefly review the concept of LAC
        in Section \ref{sec:LAC} before we propose our LACBoost and FisherBoost
        in Section \ref{sec:LACBoost}. We present the experiments in Section
        \ref{sec:exp} and conclude the paper in Section \ref{sec:con}.
        }

\section{Linear Asymmetric Classification}
\label{sec:LAC}

        Before we propose our LACBoost and FisherBoost, we 
        briefly overview the concept of LAC. 
        Wu {\em et al.} \cite{Wu2008Fast} have  proposed linear asymmetric classification
        (LAC) as a post-processing step 
        for training nodes in the cascade framework. 
        LAC is guaranteed to get an optimal solution
        under the assumption of Gaussian data distributions.

        Suppose that we have a linear classifier 
        $ f(\x) = {\bf sign}(\w^\T \x - b)$,
        if we want to find a pair of $\{ \w , b \}$ with a very
        high accuracy on the positive data $\x_1$ and a moderate accuracy on
        the negative $\x_2$, which is expressed as the following problem:
    \begin{align}
    \begin{split}
        \max_{\w \neq {\bf 0}, b}  \,  \Pr_{\x_1 \sim ( \mu_1, \bSigma_1) }
        \{ \w ^\T \x_1 \geq b \}, \,\,
        {\rm s.t.}  \, \Pr_{\x_2 \sim (\mu_2,\bSigma_2)}  
        \{ \w^\T \x_2
            \leq b \} = \lambda,
    \label{EQ:LAC}
    \end{split}
    \end{align}
            where $\x \sim (\mu,\bSigma)$ denotes 
        a symmetric distribution with mean $\mu$ and covariance $\bSigma$.
    If we prescribe $\lambda$ to $0.5$ and assume that for any $\w$, 
    $\w^\T \x_1$
    is Gaussian and $\w^\T \x_2$ is symmetric, then 
    \eqref{EQ:LAC} can be approximated by 
    \begin{equation}
        \label{EQ:LAC1}
        \max_{\w \neq \bf 0} \;\;
        \frac{  \w^\T ( \mu_1 -  \mu_2  ) } 
        {  \sqrt{  \w^\T \bSigma_1 \w  }  }.
    \end{equation}
    \eqref{EQ:LAC1} is similar to LDA's optimization problem
    \begin{equation}
        \label{EQ:LDA1}
        \max_{\w \neq \bf 0} \;\;
        \frac{  \w^\T ( \mu_1 -  \mu_2  ) } 
        {  \sqrt{  \w^\T ( \bSigma_1 + \bSigma_2 )  \w  }  }.
    \end{equation}
    \eqref{EQ:LAC1} can be solved by eigen-decomposition and a close-formed 
    solution can be derived:
\begin{equation}
            \w^\star = \bSigma_1^{-1} ( \mu_1 - \mu_2 ),
    \quad 
    b^{\star} = { \w^{\star} } ^{\T} \mu_2.
\label{EQ:LAC_SOL}
\end{equation}
On the other hand, each node in cascaded boosting classifiers has the following form:
\begin{equation}
    \label{EQ:nodeclassifier}
    f(\x) = {\bf sign}(\w^\T \H (\x) - b),
\end{equation}
We override the symbol $ \H (\x)$ here, 
which denotes the output vector of all weak classifiers over the datum $ \x $. 
We can cast each node as a linear classifier over the feature space
constructed by the binary outputs of all weak classifiers.
For each node in cascade classifier, we wish to maximize the detection
rate as high as possible, and 
meanwhile keep the false positive rate to an
moderate level  (\eg, $50.0\%$). 
That is to say,  the problem
\eqref{EQ:LAC} expresses the node learning goal. 
Therefore, we can use boosting algorithms (\eg, AdaBoost) as feature
selection methods, and then use LAC to learn a linear classifier over
those binary features chosen by boosting.
The advantage is that LAC considers the asymmetric node learning explicitly.

However, there is a precondition of LAC's validity. That 
is,  for any $\w$, $\w^\T \x_1$ is a Gaussian and $\w^\T \x_2$
is symmetric. 
In the case of boosting classifiers, $\w^\T \x_1$ and $\w^\T \x_2$ can be 
expressed as the margin of positive data and negative data.
Empirically  Wu {\em et al.} \cite{Wu2008Fast}
verified that $\w^\T \x$ is Gaussian approximately for a cascade face detector.
We discuss this issue in the experiment part in more detail.
%
%
%

\section{Constructing Boosting Algorithms from LDA and LAC}
\label{sec:LACBoost}

    In kernel methods, the original data are non\-linear\-ly 
    mapped to a feature space and 
    usually the mapping
    function $ \Phi ( \cdot ) $ is not explicitly available.
    It works through the inner product of 
    $ \Phi ( \x_i ) ^\T \Phi ( \x_j )  $. 
    In boosting \cite{Ratsch2002BoostSVM},
    the mapping function can be seen as explicitly known
    through:
    $
        \Phi ( \x ) : \x \mapsto [ h_1(\x),\dots,h_n(\x) ]. 
    $
    Let us consider the Fisher LDA case first because the solution to LDA
    will generalize to LAC straightforwardly, by looking at
    the similarity between \eqref{EQ:LAC1} and \eqref{EQ:LDA1}.

    Fisher LDA 
    maximizes the between-class variance and minimizes the within-class
    variance. In the binary-class case, we can equivalently
    rewrite \eqref{EQ:LDA1} into
    \begin{equation}
        \label{EQ:100}
        \max_\w \;\;  \frac{ ( \mu_1 - \mu_2 ) ^ 2 }
                         { \sigma_1 + \sigma_2 } 
            = 
                    \frac{   \w ^\T   \C_b \w }
                         {   \w ^\T  \C_w \w },
    \end{equation}
    where $ \C_b $ and $ \C_w $ are the between-class and within-class
    scatter matrices; $ \mu_1 $ and $ \mu_2 $ are
    the projected centers of the two classes.
    The above problem can be equivalently reformulated as 
    \begin{equation}
        \label{EQ:101}
        \min_\w \;\;  \w ^\T \C_w \w -  \theta ( \mu_1 - \mu_2  )
    \end{equation}
    for some certain constant $ \theta $ and under the assumption that
    $ \mu_1 - \mu_2 \geq 0 $.\footnote{In our face detection experiment,
    we found that this assumption could always be satisfied.}
    Now in the feature space, our data are 
    $ \Phi( \x_i ) $, $ i=1\dots m$.
    We have
    \begin{align}
        \mu_1
        &  = \frac{ 1 } { m_1 } \w^\T \sum_{y_i = 1}  \Phi(\x_i)  
           = \frac{ 1 } { m_1 } \sum_{y_i = 1} \A_{ i: } \w
        = \frac{ 1 } { m_1 } \sum_{y_i = 1} (\A \w)_i
           = \e_1 ^\T \A \w   ,
    \end{align}
    where $ \A_{ i: } $ is the $ i $-th row of $ \A$.
    \begin{align}
     \mu_2
     & = 
     \frac{ 1 } { m_2 }   \w^\T \sum_{y_i = -1}  \Phi(\x_i)
     = \frac{ 1 } { m_2 } \sum_{y_i = -1} \H_{ i: } \w
     = - \e_2 ^\T  \A \w,
    \end{align}
    Here the $ i $-th entry of $ \e_1 $ is defined as 
    $ \e_{1i} = 1/m_1 $ if $ y_i = +1 $, otherwise   
    $ \e_{1i} = 0$. Similarly 
    $ \e_{2i} = 1/m_2 $ if $ y_i = -1 $, otherwise 
    $ \e_{2i} = 0$. We also define $ \e = \e_1 + \e_2 $.
    For ease of exposition, we order the training data according to their
        labels. So
        the vector $ \e \in \Real^{m}$:
        \begin{equation}
        \e = [ 1/m_1,\cdots, 1/m_2,\cdots  ]^\T,      
            \label{EQ:e}
        \end{equation}
        and the first $ m_1$ components of $ \brho $ correspond to the
        positive training data and the remaining ones
        correspond to the $ m_2$
        negative data.  
        So we have $ \mu_1 - \mu_2 = \e^\T \brho  $, 
        $ \C_w =  {m_1 }/{ m } \cdot \bSigma_1 + {m_2 }/{ m } \cdot  \bSigma_2 $
        with
        $ \bSigma_{1,2} $ the covariance matrices. 
        By noticing that
        \[
        \w^\T \bSigma_{1,2} \w = \frac{1}{m_{1,2} ( m_{1,2} - 1 ) }
        \sum_{i>k, y_i=y_k = \pm 1}
        (\rho_i - \rho_k )^2,
        \]
        we can easily rewrite the original problem into:
        \begin{align}
            \min_{\w,\brho}
           %
           %
            \tfrac{1}{2} \brho ^\T \Q \brho - \theta \e^\T
            \brho,
            \quad {\rm s.t.} ~&\w \psd {\bf 0},
             {\bf 1}^\T \w = 1,
     {\rho}_i = ( \A \w )_i,
        i = 1,\cdots, m.
            \label{EQ:QP1}
        \end{align}
        Here
        $ \Q = \begin{bmatrix} \Q_1 & {\bf 0} \\ {\bf 0} & \Q_2  \end{bmatrix} $
        is a block matrix with
        \[
        \Q_1 = 
        \begin{bmatrix}
                \tfrac{1}{m} & -\tfrac{1}{ m (m_1-1)} & \ldots & -\tfrac{1}{m(m_1-1)} \\
                -\tfrac{1}{m(m_1-1)} & \tfrac{1}{ m } & \ldots & -\tfrac{1}{m(m_1-1)} \\
                \vdots & \vdots & \ddots & \vdots \\
                -\tfrac{1}{m(m_1-1)} & -\tfrac{1}{m (m_1-1)} & \ldots &\tfrac{1}{m } 
        \end{bmatrix},
        \]
        and $ \Q_2 $ is similarly defined by replacing $ m_1$ with $ m_2 $ in $ \Q_1$.
        \comment{
        \[
        \Q_2 = 
            \begin{bmatrix}
                \tfrac{1}{m}         & -\tfrac{1}{m(m_2-1)} & \ldots &
                -\tfrac{1}{m(m_2-1)}                                \\
                -\tfrac{1}{m(m_2-1)} & \tfrac{1}{m}         & \ldots &
                -\tfrac{1}{m(m_2-1)}                                \\
                \vdots               & \vdots & \ddots & \vdots     \\
                -\tfrac{1}{m(m_2-1)} & -\tfrac{1}{m(m_2-1)} & \ldots
                &\tfrac{1}{m} 
            \end{bmatrix}. 
        \]
        }
  Also note that we have introduced a constant $ \frac{1}{2} $ before the quadratic term
  for convenience. The normalization 
                  constraint $ { \bf 1 } ^\T \w = 1$
                  removes the scale ambiguity of $ \w $. Otherwise the problem is
                  ill-posed. 

  In the case of LAC, the covariance matrix of the negative data is not involved,
  which corresponds to the matrix $ \Q_2 $ is zero. So we can simply set 
   $ \Q = \begin{bmatrix} \Q_1 & {\bf 0} \\ {\bf 0} & \bf 0  \end{bmatrix} $ and
  \eqref{EQ:QP1} becomes the optimization problem of LAC.  
  
  At this stage, it remains unclear about how to solve the problem \eqref{EQ:QP1}
  because we do not know all the weak classifiers. 
  The number of possible weak classifiers
  could be infinite---the dimension of the 
  optimization variable $ \w $ is infinite.
    So \eqref{EQ:QP1} is a semi-infinite quadratic program (SIQP).
    We show how column generation can be used to solve this problem.
    To make column generation applicable, we need to derive a
    specific Lagrange dual of the primal
    problem.

%
%
%
\textbf{The Lagrange  dual problem}
    We now derive the Lagrange dual of the quadratic problem \eqref{EQ:QP1}.    
    Although we are only interested in the variable $ \w $, we need to
    keep the auxiliary variable $ \boldsymbol  \rho $ in order to obtain
    a meaningful dual problem. The Lagrangian of \eqref{EQ:QP1}
    is
    $ L (  
           \underbrace{ \w, \brho}_{\rm primal}, \underbrace{ \u, r }_{\rm dual}
        ) 
    = \tfrac{1}{2} \brho ^\T \Q \brho   -  \theta \e^\T \brho 
    + \u ^\T ( \brho - \A \w  ) - \q ^\T \w + r ( {\bf 1} ^\T \w - 1 )
    $ with $  \q \psd \bf 0 $.  
    $ \sup_{\u, r} \inf_{ \w, \brho } L ( \w, {\brho}, \u, r  ) $
    gives the following  Lagrange dual:
           \begin{align}
               \max_{\u, r} ~& -r - \overbrace{
                       \tfrac{1}{2} 
                      (\u - \theta \e)^\T \Q^{-1} (\u - \theta \e)
                      }^{\rm regularization},  
                      %
                      %
            {\rm \;\; s.t.} 
                     ~
                     %
                     %
                      \sum_{i=1}^m u_i \A_{i:} \nsd r {\bf 1 } ^\T. 
            \label{EQ:dual}
        \end{align}
        In our case, $ \Q $ is rank-deficient and its inverse does not exist
        (for both LDA and LAC).
        We can simply regularize $ \Q $ with $ \Q + \delta {\bf I} $ with  
        $ \delta $ a very small constant. 
        One of the  KKT optimality conditions between the dual and primal
        is
        $ \brho^\star = - \Q^{-1} ( \u ^ \star - \theta \e )$,
        which can be used to establish the connection between the dual optimum and
        the primal optimum. 
        This is obtained by the fact that 
        the gradient of $ L $ w.r.t. $ \brho $ must vanish at 
        the optimum, $ { \partial L } / { \partial \rho_i } = 0  $,
        $ \forall i = 1\cdots n $.

        Problem \eqref{EQ:dual} can be viewed as a regularized LPBoost problem.
        Compared with the hard-margin LPBoost \cite{Demiriz2002LPBoost},
        the only difference is the regularization term in the cost function.
        The duality gap between the primal \eqref{EQ:QP1} and the 
        dual \eqref{EQ:dual} is zero. In other words, the solutions of
        \eqref{EQ:QP1} and \eqref{EQ:dual} coincide. 
        Instead of solving \eqref{EQ:QP1} directly, one calculates the
        most violated constraint in \eqref{EQ:dual} iteratively for
        the current solution and adds this constraint to the
        optimization problem.  In theory, any column that violates
        dual feasibility can be added.  To speed up the convergence,
        we add the most violated constraint by solving the following
        problem:
      \begin{equation}
          h' ( \cdot ) =  {\rm argmax}_{h( \cdot ) } ~ 
            \sum_{i=1}^m u_i y_i h ( \x_i).
         \label{EQ:pickweak}
      \end{equation}
        This  is exactly the same as the one that standard AdaBoost
        and LPBoost use for producing the best weak classifier. That
        is to say, to find the weak classifier that has minimum weighted
        training error.  We summarize the LACBoost/FisherBoost
        algorithm in
        Algorithm~\ref{alg:QPCG}.
        By simply changing  $ \Q_2 $, Algorithm~\ref{alg:QPCG} can be used to
        train either LACBoost or FisherBoost.
        Note that to obtain an actual strong classifier,
        one may need to include an offset $ b $, {\em i.e.} the final classifier
        is $ \sum_{j=1}^n h_j (\x) - b $ because from the cost function
        of our algorithm \eqref{EQ:101}, we can see that the cost function itself
        does not minimize any classification error. It only finds a projection
        direction in which the data can be maximally separated. A simple line
        search can find an optimal $ b $.  
        Moreover, when training a cascade, we need to tune this offset anyway
        as shown in \eqref{EQ:nodeclassifier}.

        The convergence of Algorithm~\ref{alg:QPCG} is guaranteed by
        general column generation or cutting-plane algorithms, which
        is easy
        to establish.  When a new $ h'(\cdot) $ that violates dual
        feasibility is added, the new optimal value of the dual
        problem (maximization) would decrease.  Accordingly, the
        optimal value of its primal problem decreases too because they
        have the same optimal value due to zero duality gap. Moreover
        the primal cost function is convex, therefore in the end it
        converges to the global minimum.

   \linesnumbered\SetVline
   \begin{algorithm}[t]
   \caption{Column generation for QP.}   
   \centering
   \begin{minipage}[]{0.91\linewidth}
   \KwIn{Labeled training data $(\x_i, y_i), i = 1\cdots m$;
         termination threshold $ \varepsilon > 0$;
         regularization
         parameter $ \theta $; maximum number of iterations
         $ n_{\rm max}$.
    }
       { {\bf Initialization}:
            $ m = 0 $;
            $ \w = {\bf 0} $;
            and $ u_i = \frac{1}{ m }$, $ i = 1$$\cdots$$m$. 
   }

   \For{ $ \mathrm{iteration} = 1 : n_\mathrm{max}$}
   {
     %
     %
     \ADot
         Check for the optimality: \\
         {\bf if}{ $ \mathrm{iteration}  > 1 $ \text{ and } $
                    \sum_{ i=1 }^m  u_i y_i h' ( \x_i )  
                           < r + \varepsilon $},
                  \\
                  { \bf then}
                  \\
                  $~ ~ ~$ break;  and the problem is solved; 
        
     \ADot
         Add $  h'(\cdot) $ to the restricted master problem, which
         corresponds to a new constraint in the dual;
     %
     %
     %

      \ADot  
         Solve the dual problem \eqref{EQ:dual}
         (or the primal problem \eqref{EQ:QP1}) 
         and update $ r $ and
         $ u_i$ ($ i = 1\cdots m$). 
%
%

      \ADot  
         Increment the number of weak classifiers
             $n = n + 1$.  
%
%
%
   }
   \KwOut{
         %
         The selected features are $ h_1, h_2, \dots, h_n $.
         The final strong classifier is:
         $ F ( \x ) = \textstyle \sum_{j=1}^{ n } w_j h_j( \x ) - b $.
         Here the offset $ b $ can be learned by a simple search. 
     
   }
   \end{minipage}
   \label{alg:QPCG}
   \end{algorithm}

    At each iteration of column generation,
    in theory, we can  solve either the dual \eqref{EQ:dual} 
    or the primal problem \eqref{EQ:QP1}. 
    However, 
    in practice, it could be  much faster to solve the primal problem because
 (i) Generally,
    the primal problem has a smaller size, hence faster to solve.
    The number of variables of \eqref{EQ:dual} is $ m $ at each iteration,
    while the number of variables is the number of iterations   
    for the primal problem. 
    For example, in Viola-Jones' face detection framework, 
    the number of training data $ m = 
    10,000 $ and  $ n_{\rm max} = 200 $. In other words, the  
    primal problem has at most $ 200 $ variables in this case;
   (ii)
        The dual problem is a standard QP problem. It has no special structure
        to exploit. As we will show, the primal problem belongs to
        a special class of problems and
        can be efficiently 
        solved using entropic/exponentiated 
        gradient descent (EG) \cite{Beck03Mirror,Globerson07Exp}. 
        A fast QP solver is extremely important for training a 
        object detector because we need to the solve a few thousand 
        QP problems. 

    We can recover both of the dual variables 
    $ \u^\star, r^\star $ easily from 
    the primal variable $ \w^\star $:
    \begin{align}
        \u^\star &=  - \Q\brho^\star  + \theta \e; \label{EQ:KA}\\
         r^\star  &=   \max_{ j = 1 \dots n }  
         \bigl\{ \textstyle \sum_{i=1}^m u_i^\star \A_{ij} \bigr\}.
         \label{EQ:KAb}
    \end{align}
    The second equation is obtained by the fact that 
     in the dual problem's  constraints, at optimum,
    there must exist at least one  $ u_i^\star$ 
    such that the equality holds. That is to say,
    $ r^\star $ is the largest {\em edge}
    over all weak classifiers.

    We give a brief introduction to the EG algorithm before we proceed. 
    Let us first define the unit simplex 
    $ \Delta_n =  \{ 
    \w \in \Real^n  :  {\bf 1 } ^ \T \w = 1, \w \psd {\bf 0 }
    \} $. 
    EG efficiently solves the convex optimization problem
    \begin{equation}
        \label{EQ:EG1}
        \min_\w \,\,\, f(\w), \,
        {\rm s.t.} \,\, \w \in \Delta_n,  
    \end{equation}
    under the assumption that the objective function $ f(\cdot) $
    is a convex Lipschitz continuous function with Lipschitz
    constant $ L_f $ w.r.t. a fixed given norm $ \lVert \cdot \rVert$.
    The mathematical definition of $ L_f $ is that
    $  | f(\w) -f (\z) |  \leq L_f   \lVert  \x - \z  \rVert$ holds
    for any $ \x, \z $ in the domain of $ f(\cdot)$.
    The EG algorithm is very simple:
    \begin{enumerate}
    \item
        Initialize with $\w^0 \in \text{the interior of }  \Delta_n$;
    \item
        Generate the sequence $ \{ \w^k \} $, $ k=1,2,\cdots$
        with:
        \begin{equation}
            \label{EQ:EQ2}
            \w^k_j = \frac{ \w^{k-1}_j \exp [ - \tau_k f'_j ( \w^{k-1} ) ]  } 
            { \sum_{j=1}^n  \w^{k-1}_j \exp [ - \tau_k f'_j ( \w^{k-1} ) ] }. 
        \end{equation}
        Here $ \tau_k $ is the step-size. 
        $ f'( \w ) = [ f_1'(\w), \dots,  f_n'(\w) ] ^\T $
        is the gradient of $ f(\cdot) $;
    \item
        Stop if some stopping criteria are met.
    \end{enumerate}
    The learning step-size can be determined by 
        $    \tau_k = \frac{ \sqrt{ 2\log n } } { L_f }
                     \frac{1}{ \sqrt{ k } },
        $
    following \cite{Beck03Mirror}.
    In \cite{Globerson07Exp}, the authors have 
    used a simpler strategy to set the learning rate. 

    EG is a very useful tool for solving large-scale 
    convex minimization problems over the unit simplex. 
    Compared with standard QP solvers like Mosek 
    \cite{Mosek}, EG is much faster. EG makes it possible 
    to train a detector using almost the same amount of time
    as using standard AdaBoost as the majority of time is
    spent on weak classifier training and bootstrapping.

    In the case that $ m_1 \gg 1 $, 
    \[
            \Q_1 =
            \frac{1}{m}
    \begin{bmatrix}
        1   & -\tfrac{1}{  m_1-1} & \ldots & -\tfrac{1}{  m_1-1 } \\
        -\tfrac{1}{ m_1-1 } & 1 & \ldots & -\tfrac{1}{ m_1-1} \\
        \vdots & \vdots & \ddots & \vdots \\
        -\tfrac{1}{ m_1-1 } & -\tfrac{1}{ m_1-1 } & \ldots &  1 
    \end{bmatrix} \approx  \frac{1}{m} \bf I.
    \]
    Similarly, for LDA, $ \Q_2 \approx  \frac{1}{m} \bf I$
    when $ m_2 \gg 1 $. Hence,
    \begin{equation}
     \label{EQ:Q10}    
      \Q \approx 
     \begin{cases}
          \frac{1}{m} \bf I;    & \text{for Fisher LDA},    \\
           \frac{1}{m} \begin{bmatrix}
                                          {\bf I} & {\bf 0} \\
                                          {\bf 0} & {\bf 0}
                                 \end{bmatrix},
                                & \text{for LAC}.
    \end{cases}
    \end{equation}
    Therefore, the problems involved can be simplified when $ m_1 \gg 1 $ and
    $ m_2 \gg 1 $ hold.
    The primal problem \eqref{EQ:QP1} equals
     \begin{align}
            \min_{\w,\brho} ~& \tfrac{1}{2} \w ^\T ( \A^\T \Q \A) \w  
            - ( \theta \e ^\T
            \A ) \w,
            %
            %
            \;\;
            {\rm s.t.}
            \;
            %
            %
            \w \in \Delta_n.
            \label{EQ:QP2}
        \end{align}
        We can efficiently solve \eqref{EQ:QP2} 
        using the EG method. 
        In EG there is an important parameter $ L_f $, which is
        used to determine the step-size. 
        $ L_f $ can be determined by the $\ell_\infty $-norm of $ | f' (\w) | $.
        In our case $ f' (\w) $ is a linear function, which is trivial to compute.
        The convergence of EG is guaranteed; see \cite{Beck03Mirror} for details.
        
        In summary, when using EG to solve the primal problem, 
        Line $ 5 $ of Algorithm~\ref{alg:QPCG} is:        
        
        \ADot 
        {\em Solve the primal problem \eqref{EQ:QP2} using EG, and update 
        the dual variables $ \u $ with \eqref{EQ:KA}, and $ r $ with \eqref{EQ:KAb}.
        }

        \comment{
        %
        FIXME for the journal version
        %
         In our case \eqref{EQ:QP2},
         $ L_f $ can be set to $ |  \A ^\T \A  |_\infty + \theta m \norm[1]{ \e ^\T \A }$
         with $ |  \A ^\T \A  |_\infty $
        the maximum magnitude of the matrix $ \A ^\T \A $.
        $ L_f $ is the upper bound of the $ \ell_1$-norm of the cost function's gradient:
        $  \norm [1]{  \w^\T (\A^\T \A ) + \theta m \e^\T \A    } \leq L_f $. 
        With the triangle inequality, we have
        $  \norm [1]{  \w^\T (\A^\T \A ) - \theta m \e^\T \A    } $ $
        \leq \norm[1]{ \w^\T (\A^\T \A )  } $ $ + \theta m \norm[1]{ \e^\T \A } $ $
        \leq |  \A ^\T \A  |_\infty $ $ + \theta m \norm[1]{ \e ^\T \A }$.

        The following theorem ensures the convergence of the EG algorithm 
        for our problem.
        \begin{theorem}
            Under the assumption that the learning step-size
            satisfies $ 0 < \tau_k  <  \frac{1} { n | \A^\T \A |_\infty  }  $,
            then we have
            $ f( \w^\star  ) \leq f (\w ^{ k } ) \leq f (\w ^\star ) 
            + \frac{1} { \tau_k ( k -1 ) } {\rm KL} [  \w^\star \Vert \w^0   ] $,
            where $ f(\cdot) $ denotes the objective function in \eqref{EQ:QP2};
            and $  | \X |_\infty  $ denotes the maximum magnitude element of 
            the matrix $ \X $; $ {\rm KL} ( \u \Vert \v )$ computes the 
            Kullback–Leibler divergence of $ \u, \v \in \Delta_n $.
        \end{theorem}
        The proof follows the proof of Theorem 1 in \cite{Globerson07Exp}. 
        }


\section{Applications to Face Detection}

%
%
\begin{figure}[t!]
    \centering
        \includegraphics[width=.45\textwidth]{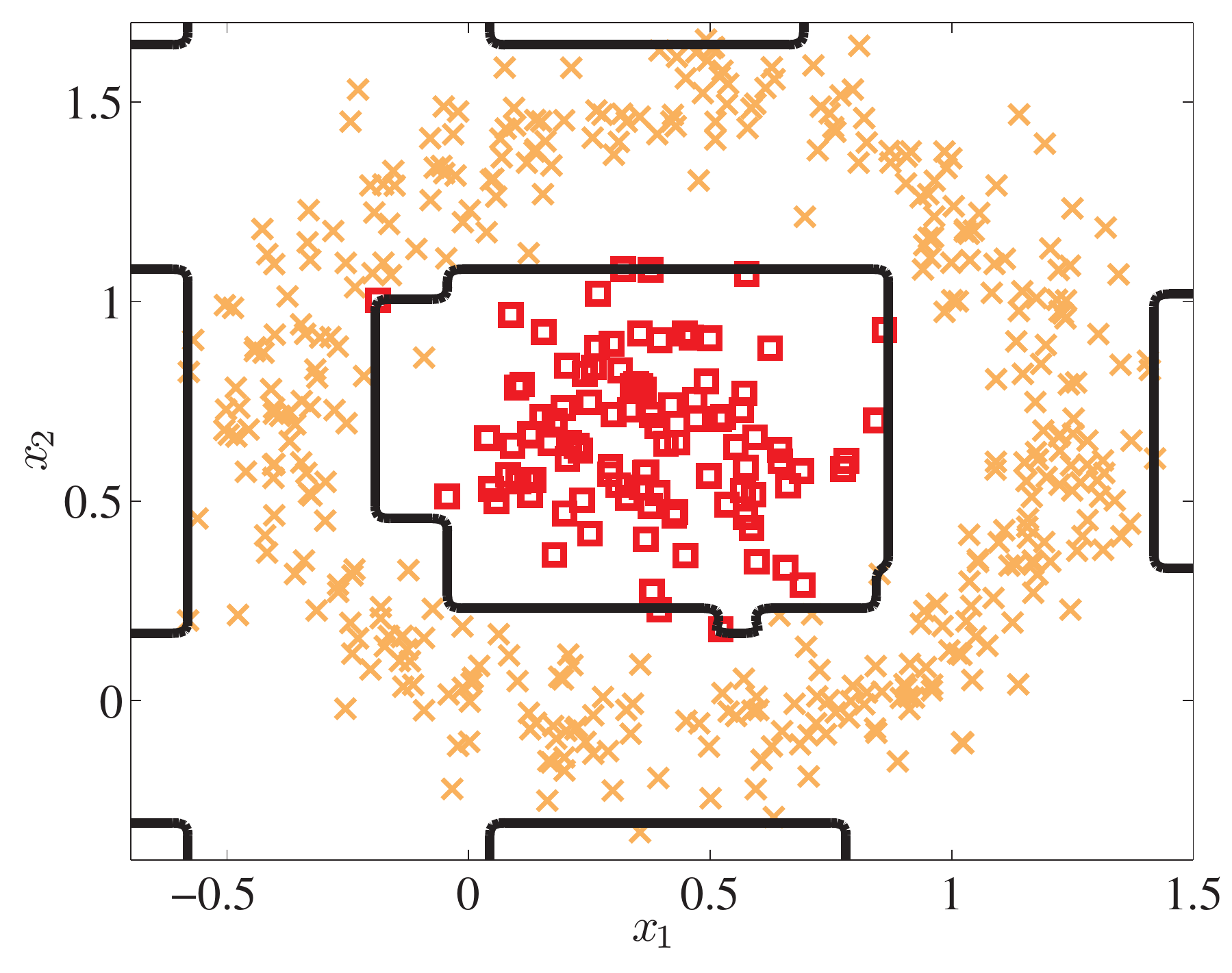}
        \includegraphics[width=.45\textwidth]{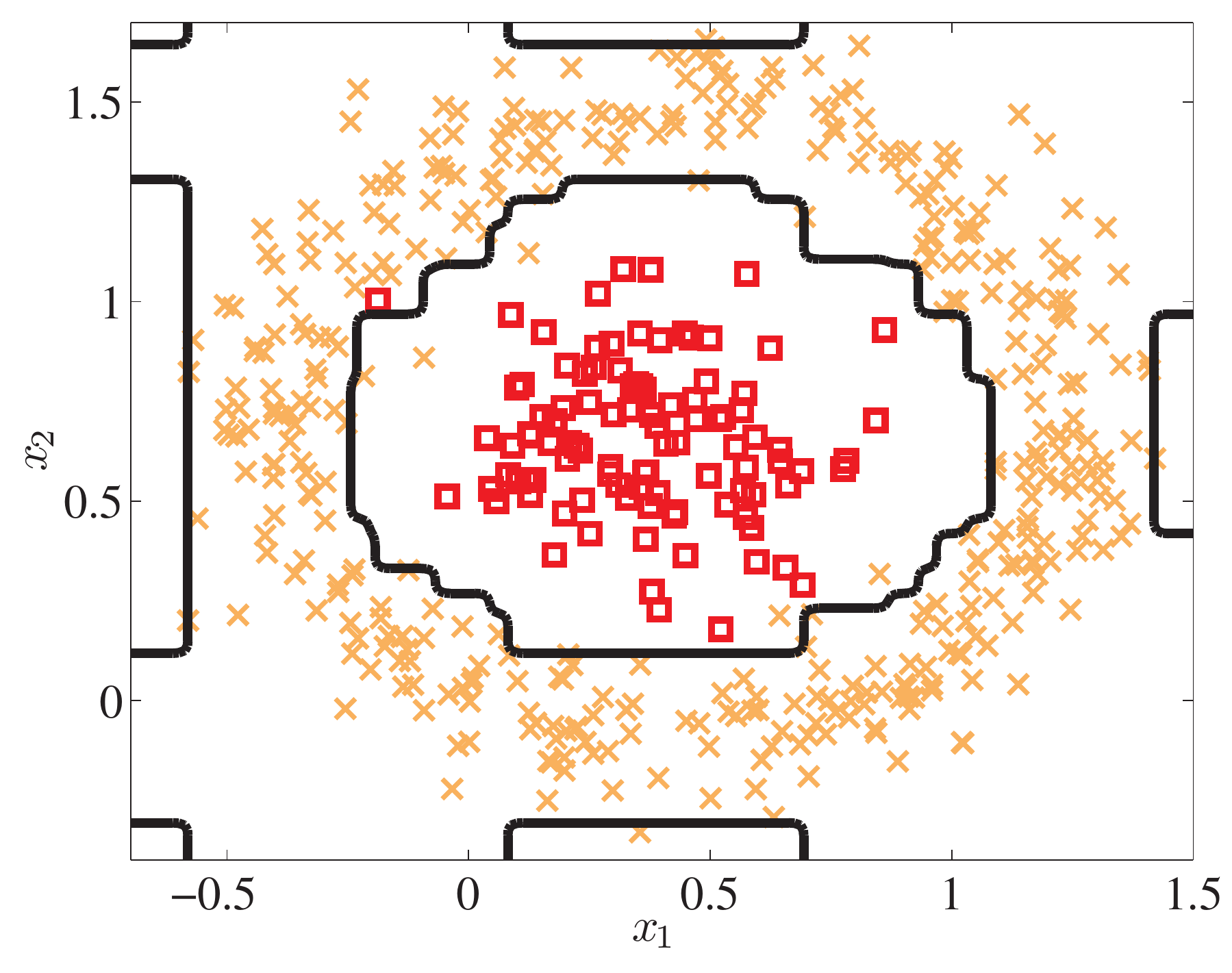}
    \caption{Decision boundaries of 
    AdaBoost (left) and FisherBoost (right) on $2$D artificial data
    (positive data represented by $ \square $'s and negative data
    by $\times$'s). Weak classifiers are decision stumps.
    In this case, FisherBoost intends to correctly classify more
    positive data in this case. 
    }
    \label{fig:toy}
\end{figure}

    First, let us show a simple example on a synthetic dataset 
    (more negative data than positive data)
    to illustrate
    the difference between FisherBoost and AdaBoost.
    Fig. \ref{fig:toy} demonstrates the subtle difference of the classification 
    boundaries obtained by AdaBoost and FisherBoost. 
    We can see that
    FisherBoost seems to focus more on correctly classifying positive data points.
    This might be due to the fact that AdaBoost only optimizes the overall 
    classification accuracy. 
    This finding is consistent with the result in \cite{Paisitkriangkrai2009CVPR}.
    
\begin{figure}[t]
    \centering
        \includegraphics[width=.32\textwidth]{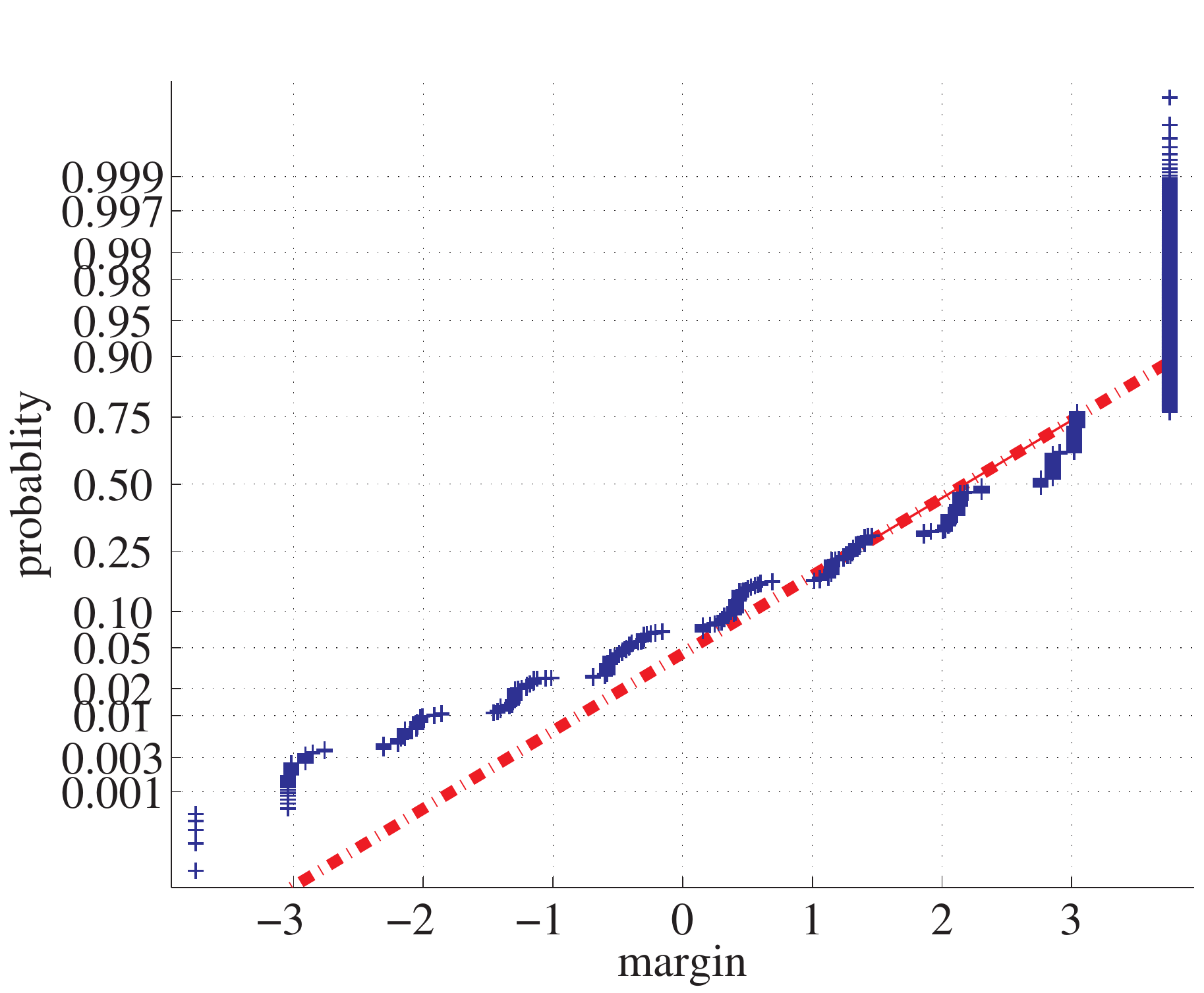}
        \includegraphics[width=.32\textwidth]{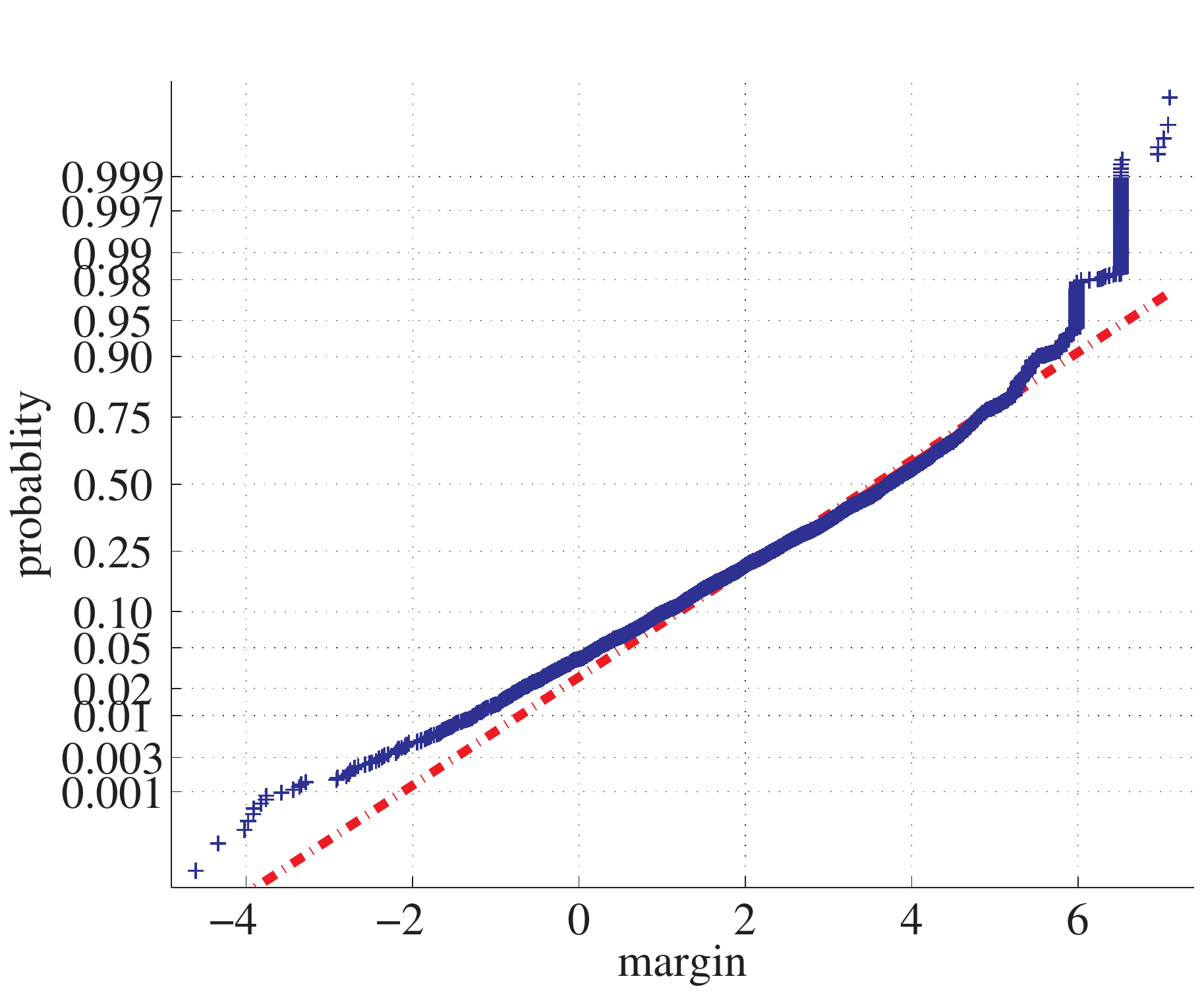}
        \includegraphics[width=.32\textwidth]{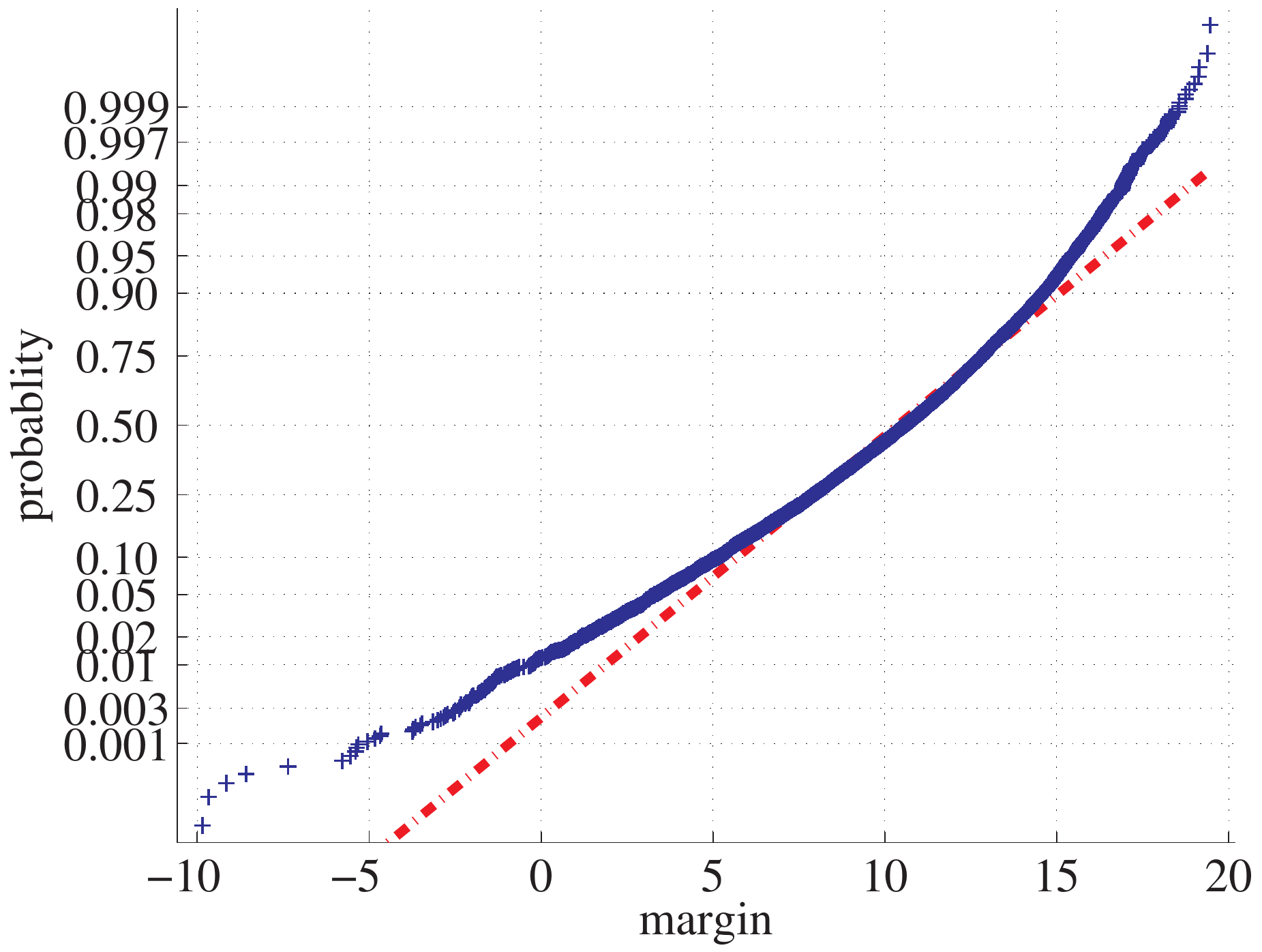}
    \caption{Normality test (normal probability plot)
    for the face data's margin distribution of nodes $1$, $2$, $3$.
    The $ 3 $ nodes contains $ 7 $, $ 22 $, $ 52 $ weak classifiers respectively.
    Curves close to a straight line mean close to a Gaussian.
    }
    \label{fig:normplot}
\end{figure}

%
%
\textbf{Face detection}
    In this section, we compare our algorithm with other 
    state-of-art face detectors.
    We first show some results about the validity 
    of LAC (or Fisher LDA) post-processing for improving 
    node learning in object detection. 
    Fig. \ref{fig:normplot} illustrates the normal probability plot of 
    margins of positive training
    data, for the first three nodes in the multi-exit with LAC cascade. 
    Clearly, the larger
    number of weak classifiers being used,
    the more closely the margin follows Gaussian distribution.
    In other words, 
    LAC may achieve a better performance if a larger number of weak classifiers
    are used.  The performance could be poor with too fewer weak classifiers.
    The same statement applies to Fisher LDA, and LACBoost, FisherBoost, too. 
    Therefore, we do not
    apply LAC/LDA in the first eight nodes because the margin distribution could be
    far from a Gaussian distribution.
    Because the late nodes of a multi-exit cascade contain more weak classifiers, 
    we conjecture that the multi-exit cascade 
    might meet the Gaussianity requirement better. We have compared 
    multi-exit cascades with LDA/LAC post-processing against 
    standard cascades with LDA/LAC post-processing in \cite{Wu2008Fast}
    and slightly improved performances were obtained.

    Six methods are evaluated with  the multi-exit cascade framework
    \cite{pham08multi},
    which are AdaBoost with LAC post-processing,
    or LDA post-processing,
    AsymBoost with LAC or LDA post-processing \cite{Wu2008Fast}, and our
    FisherBoost,
    LACBoost. We have also implemented Viola-Jones'
    face detector as the baseline \cite{Viola2004Robust}.
    As in \cite{Viola2004Robust}, five  basic types of Haar-like features 
    are calculated, which makes up of a $162, 336$ dimensional over-complete
    feature set on an image of $24 \times 24$ pixels.
%
%
%
%
    To speed up the weak classifier training, as in \cite{Wu2008Fast},
    we  uniformly sample $10\%$ of features for
    training weak classifiers (decision stumps). 
    The training data are $9,832$ mirrored $24 \times 24$ face images 
    ($5,000$ for training and $4,832$ for validation) and $7,323$ large background images,
    which are the same as in \cite{Wu2008Fast}.

    Multi-exit cascades with $22$ exits and $2,923$ weak classifiers are
    trained with various methods. 
    For fair comparisons, we have used the same cascade structure and 
    same number of weak classifiers for all the compared learning methods.
    The indexes of exits are pre-set to simplify the training
    procedure.
%
%
%
%
%
%
%
      For our FisherBoost and LACBoost, we have an important parameter  
      $\theta$, which is chosen from 
      $\{
      \frac{1}{10}, 
      \frac{1}{12},
      \frac{1}{15},
      $
      $
      \frac{1}{20}, 
      \frac{1}{25},
      \frac{1}{30},
      $
      $
      \frac{1}{40},
      \frac{1}{50} 
      \}$.
      We have not carefully tuned this parameter using cross-validation.
      Instead, we train a $10$-node cascade for each candidate $ \theta$, 
      and choose the one with the best {\em training} 
      accuracy.\footnote{ To train a complete $22$-node cascade
      and choose the best $ \theta $
      on cross-validation data may give better detection rates.} 
     At each exit, negative examples misclassified by current cascade are
     discarded, and new negative examples are bootstrapped from the background
     images pool.  
     Totally, billions of negative examples are extracted from the pool.
        The positive training data and validation data keep unchanged  during the
        training process.

Our experiments are performed on a workstation with $8$ Intel Xeon
E$5520$ CPUs and $32$GB RAM.
It takes about $3$ hours  to train the multi-exit cascade with AdaBoost or AsymBoost.
For FisherBoost and LACBoost, it takes less than $ 4 $ hours to train
a complete multi-exit cascade.\footnote{Our implementation is in C++ and 
                                        only the weak classifier
                                        training part is parallelized using OpenMP. 
}
In other words, 
%
%
%
%
%
    our EG algorithm takes less than $ 1 $ hour for  
    solving the primal QP problem (we need to solve a QP at each iteration).
    A rough estimation of the computational complexity is as follows. 
    Suppose that the number of training
    examples is $  m $, number of weak classifiers is $ n $,
%
%
%
    At each iteration of the cascade training,
    the complexity for solving the primal QP using EG is
    $  O( m n  + k n^2) $ with $ k $ the iterations 
    needed for EQ's convergence.
    The complexity  for  training the weak classifier  is
    $ O( m  d ) $ with $d$ the number of all Haar-feature patterns.
    In our experiment, $ m = 10,000 $,
    $ n \approx 2900 $,
    $d = 160,000$,
    $ k < 500 $.
So the majority of the training computation is on the weak classifier training.

    We have also experimentally observed the speedup of EG against standard QP solvers. 
    We solve the primal QP defined by \eqref{EQ:QP2} using EG and Mosek \cite{Mosek}.
    The QP's size is $ 1,000 $ variables. 
    With the same accuracy tolerance (Mosek's primal-dual gap is set to $ 10^{-7}$
    and EG's convergence tolerance is also set to $ 10^{-7}$), 
    Mosek takes $1.22 $ seconds and EG is
    $ 0.0541 $ seconds. So EG is about $ 20 $ times faster. 
    Moreover,  at iteration $ n + 1 $ of training the cascade,
    EG can take advantage of the last iteration's solution
    by starting EG from a small perturbation of the previous solution. 
    Such a warm-start gains a $ 5 $ to $ 10\times $ speedup in our experiment,
    while there is no off-the-shelf warm-start QP solvers available yet.

    We evaluate the detection performance on the MIT+CMU frontal
    face test set. 
%
%
%
%
%
Two performance metrics are used here: each node and the entire cascade.
The node metric is how well the classifiers meet the node learning objective.
The node metric provides useful information about 
the capability of each method to achieve the node learning goal.
The cascade metric uses the receiver operating characteristic (ROC)
to compare the entire cascade's peformance.
Multiple issues have impacts on the cascade's performance: classifiers,
the cascade structure,  bootstrapping {\em etc}.

\begin{figure}[t]
    \begin{center}
        \includegraphics[width=.45\textwidth]{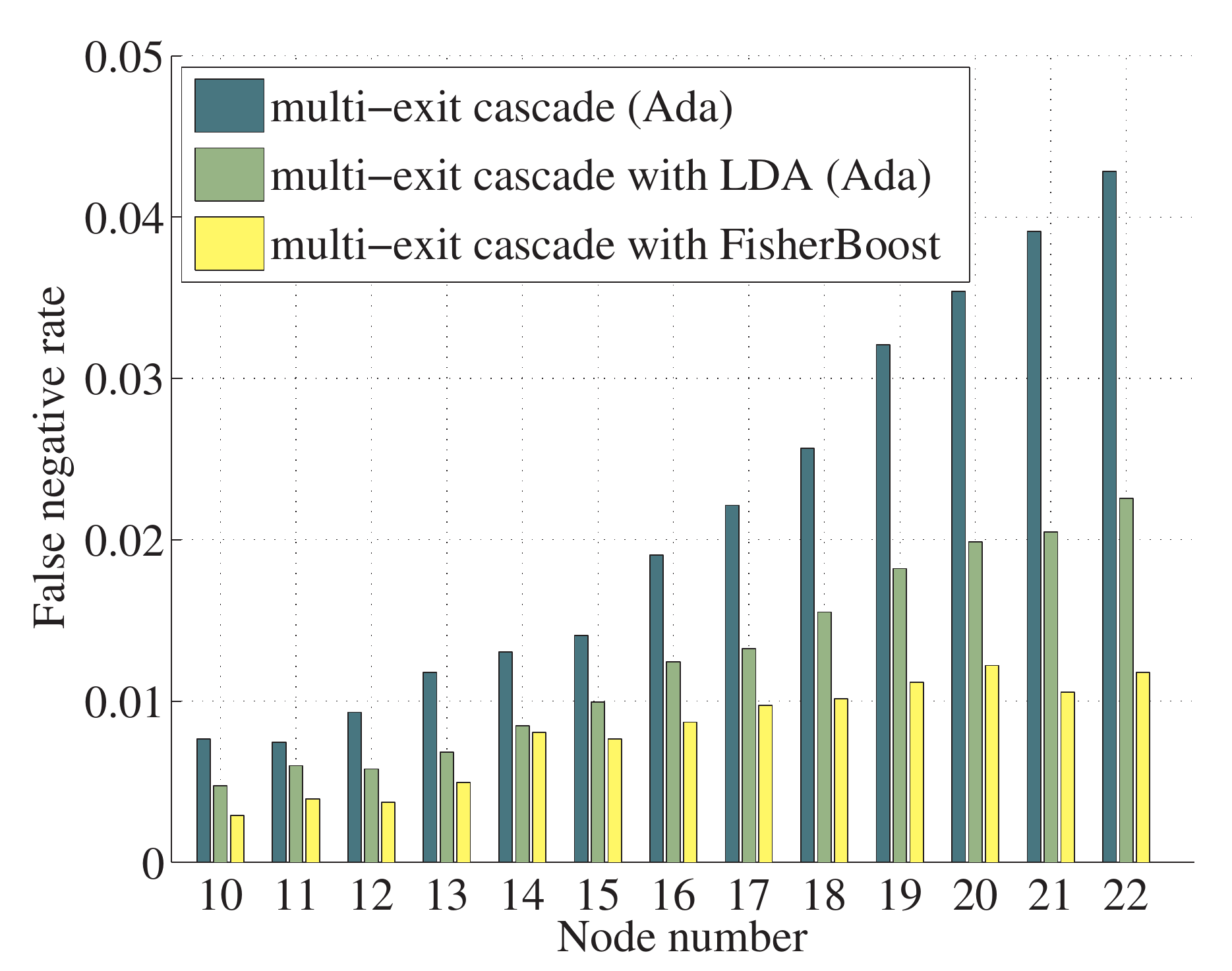}
        \includegraphics[width=.45\textwidth]{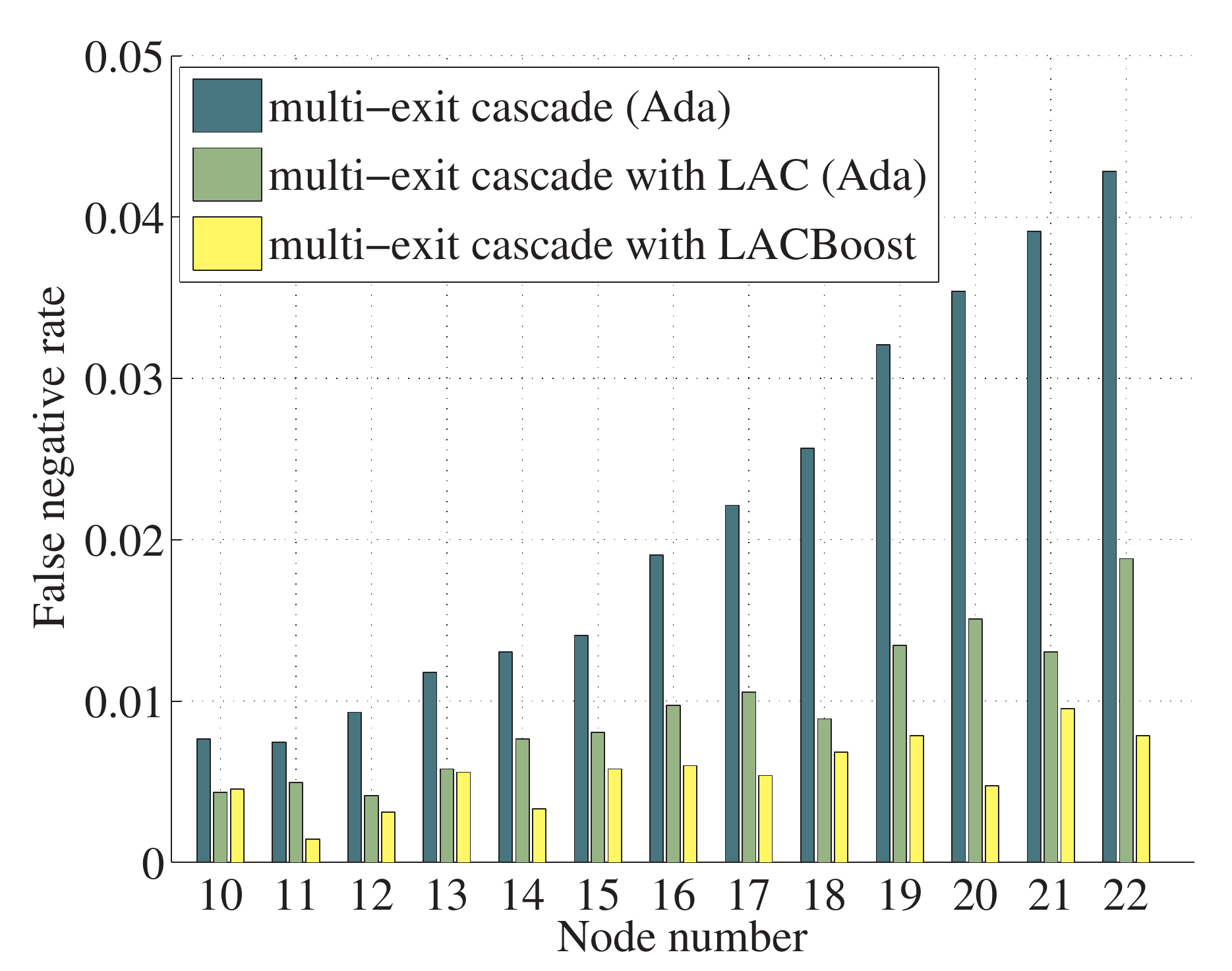}
        \includegraphics[width=.45\textwidth]{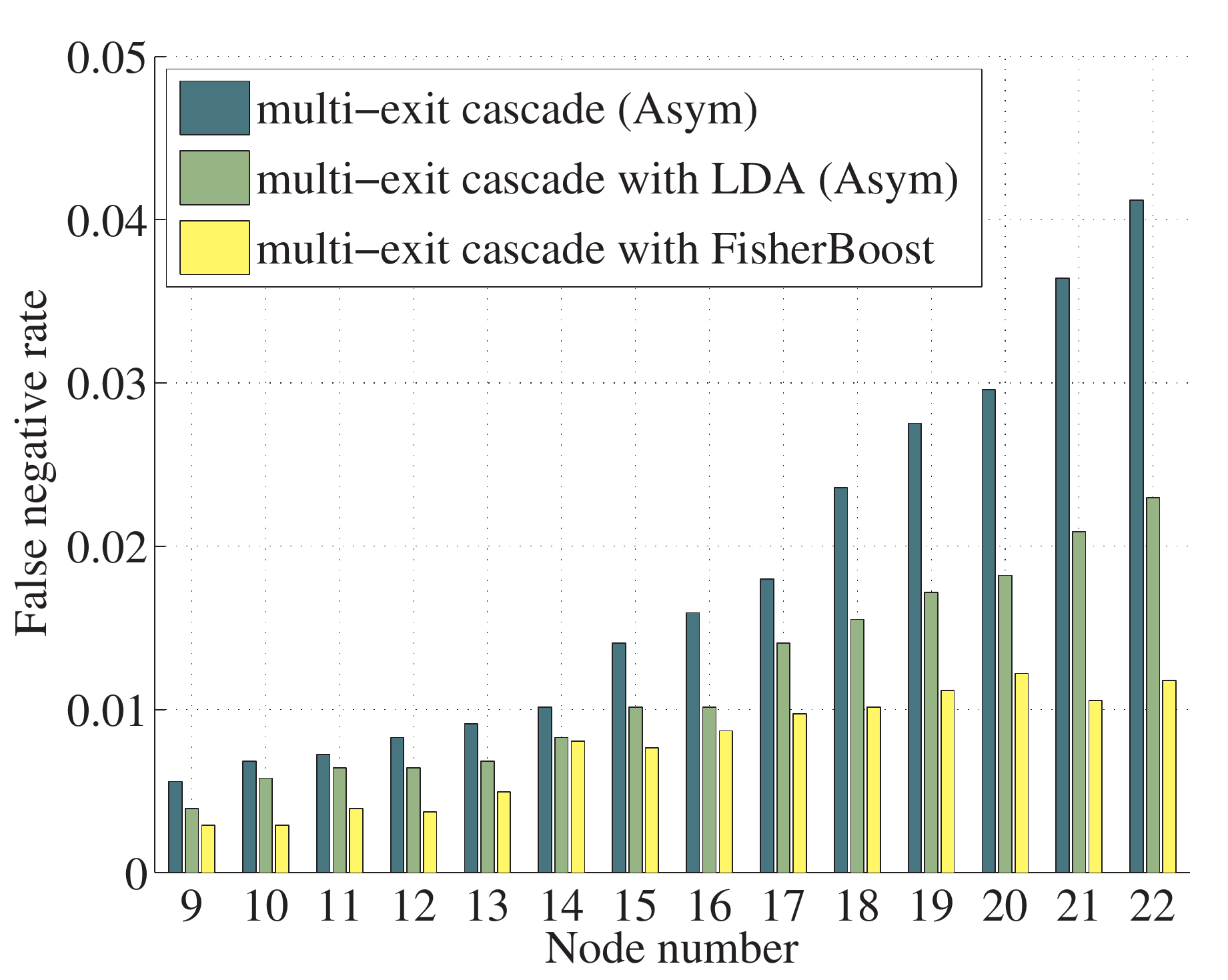}
        \includegraphics[width=.45\textwidth]{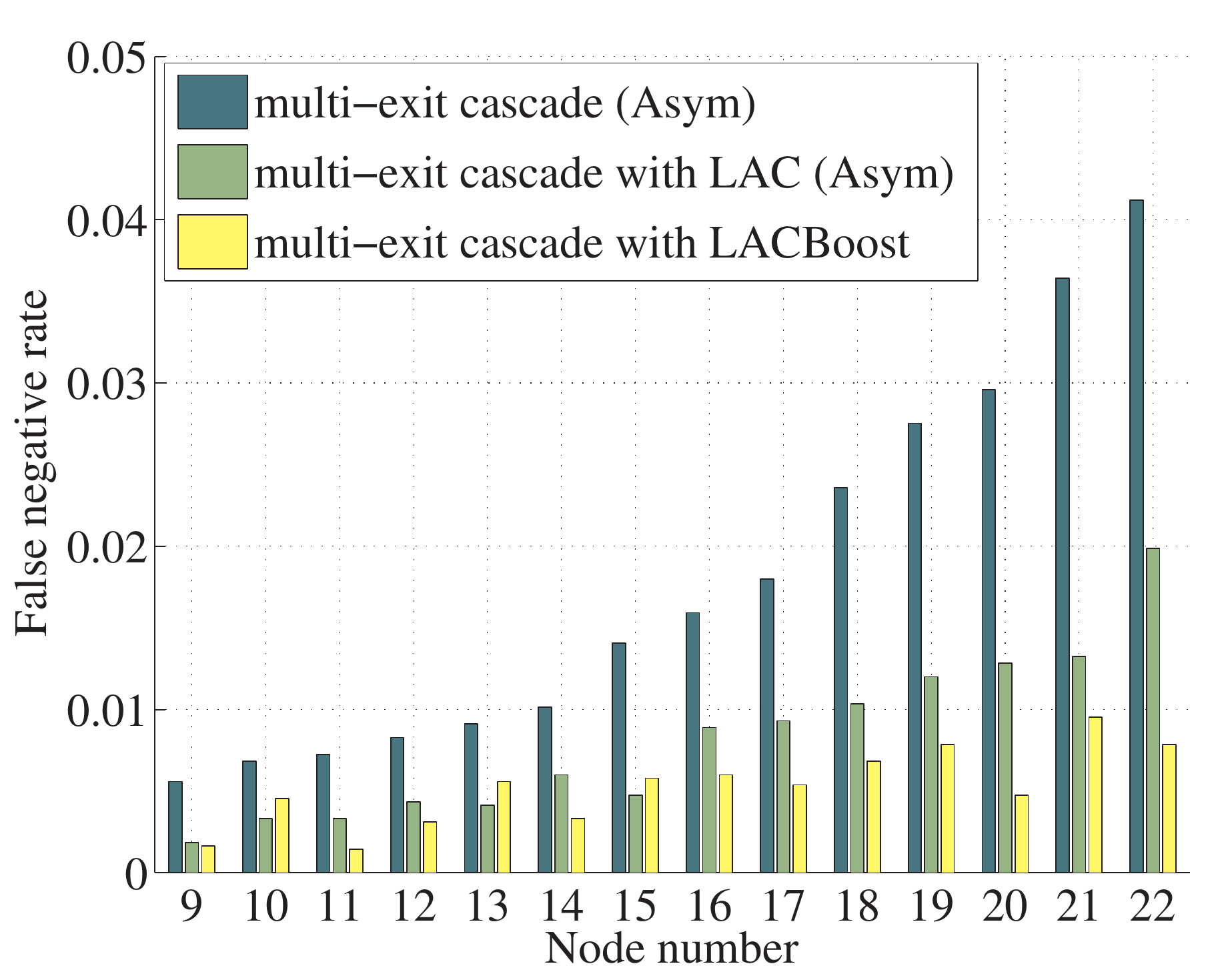}
    \end{center}
    \caption{Node performances on the validation data. 
       ``Ada'' means that features are selected using AdaBoost;
    ``Asym'' means that features are selected using AsymBoost.
    }
    \label{fig:node1}
\end{figure}

We show the node comparison results  in Fig. \ref{fig:node1}.
The node performances between FisherBoost and LACBoost
are very similar. From Fig.~\ref{fig:node1}, as reported in \cite{Wu2008Fast},
LDA or LAC post-processing can considerably reduce the
false negative rates. 
As expected, our proposed FisherBoost and LACBoost can further reduce the false
negative rates significantly. 
This verifies the advantage of selecting features with the node learning goal 
being considered. 
%
%
%

\begin{figure}[t]
    \begin{center}
        \includegraphics[width=.45\textwidth]{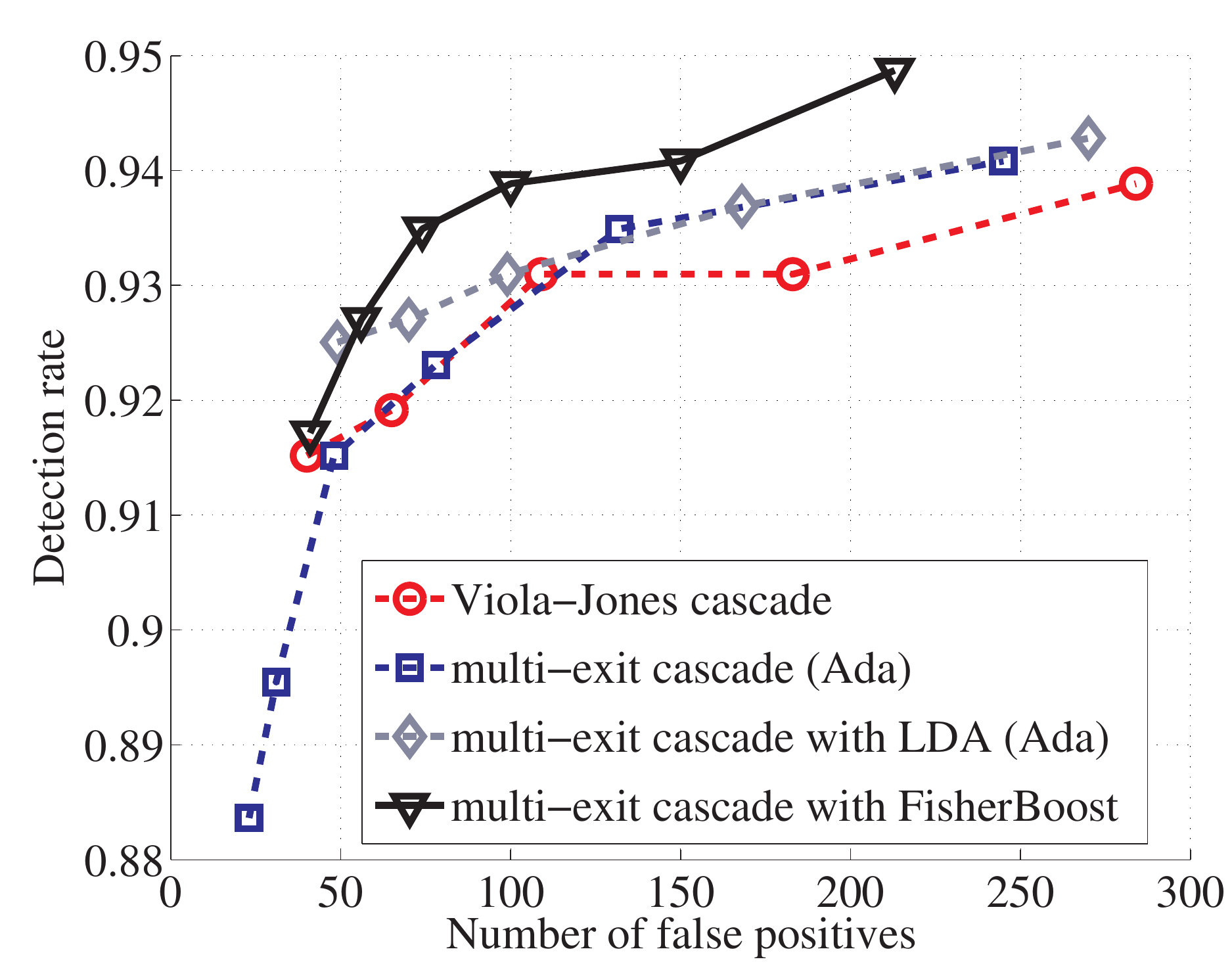}
        \includegraphics[width=.45\textwidth]{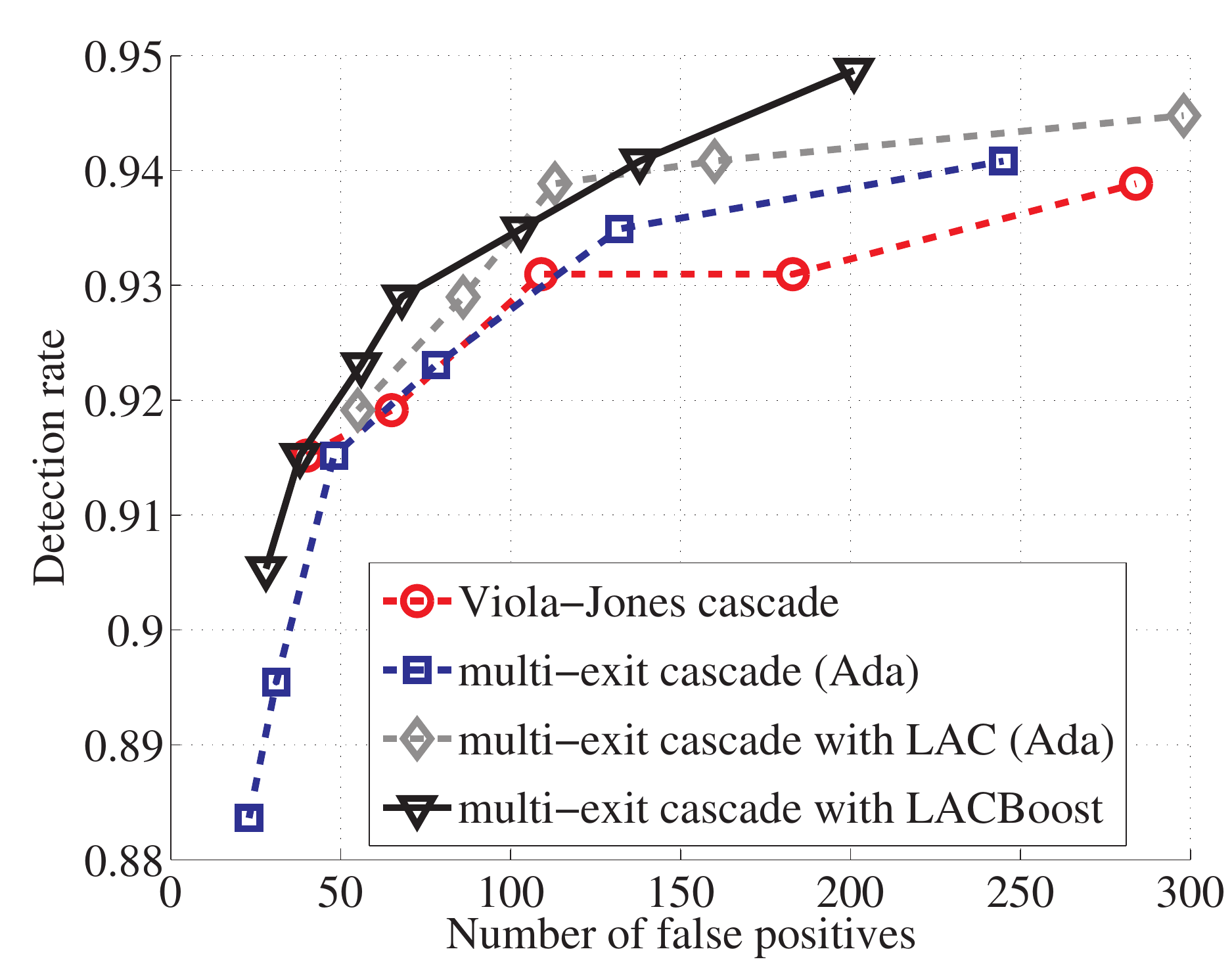}
        \includegraphics[width=.45\textwidth]{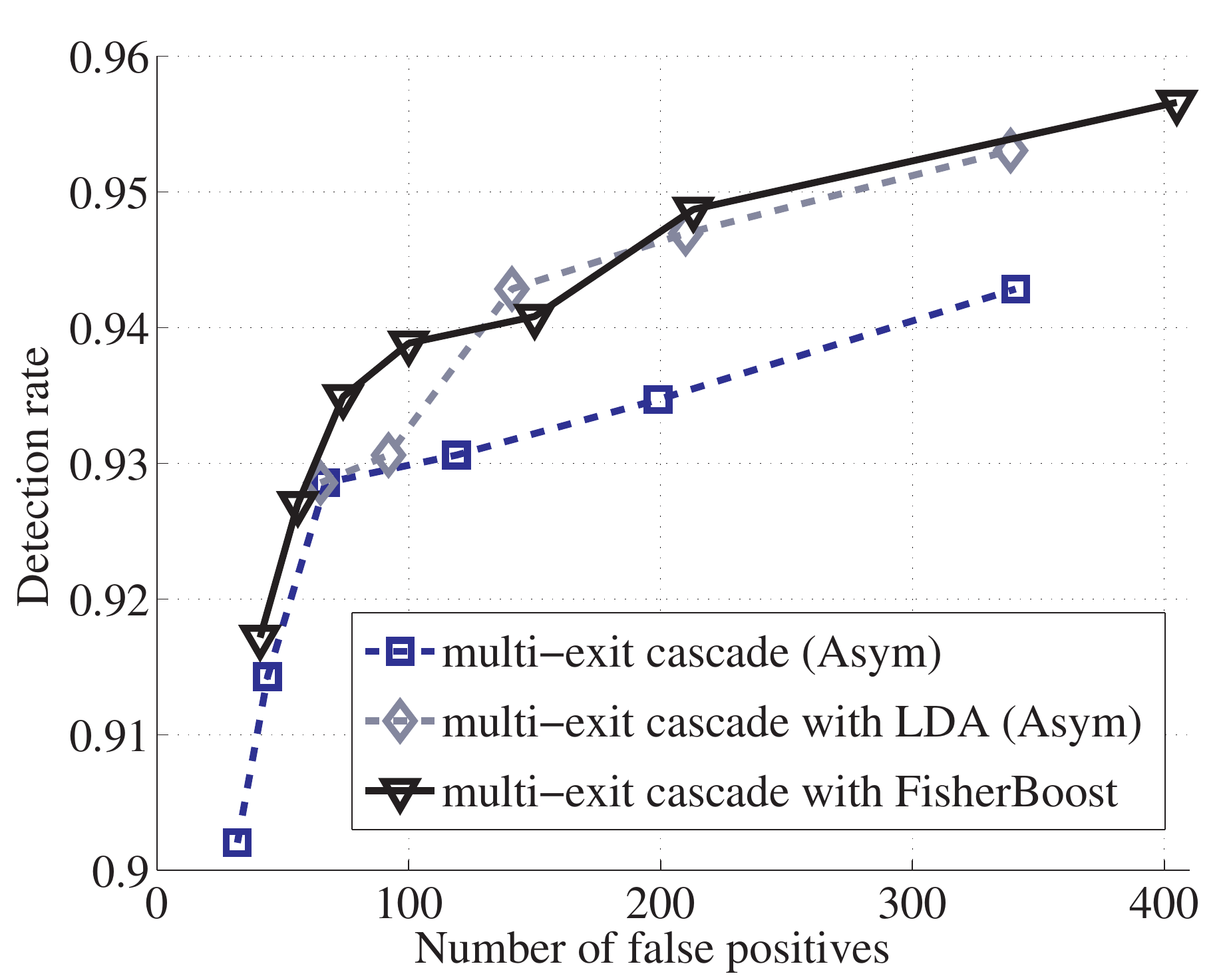}
        \includegraphics[width=.45\textwidth]{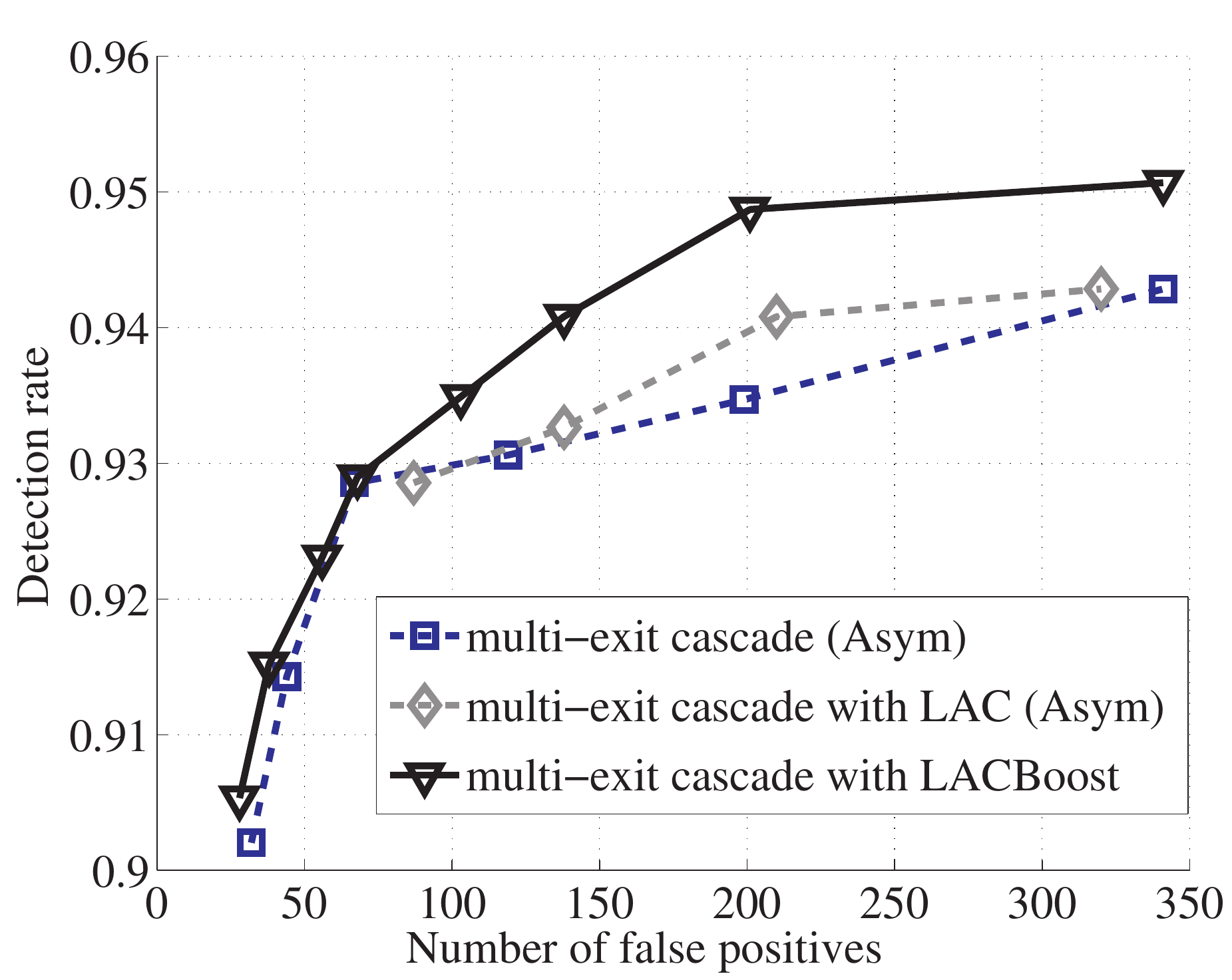}
    \end{center}
    \caption{Cascade performances using ROC curves 
    (number of false positives versus detection rate) on the MIT+CMU test data.
    ``Ada'' means that features are selected using AdaBoost. Viola-Jones cascade
    is the method in
    \cite{Viola2004Robust}.
      ``Asym'' means that features are selected using AsymBoost.
    }
    \label{fig:ROC1}
\end{figure}

From the ROC curves in Fig.~\ref{fig:ROC1}, we can see that FisherBoost and LACBoost
    outperform all the other methods.
    In contrast to the results of the detection rate for each node, 
    LACBoost is slightly worse than FisherBoost in some cases.
    That might be due to that many factors have impacts on the final result of
    detection.
%
  %
  LAC makes the assumption of Gaussianity and symmetry data distributions,
    which may not hold well in the early nodes. 
    This could explain why
    LACBoost does not always perform the best.
    Wu {\em et al.} have observed the same phenomenon that LAC post-processing
    does not outperform LDA post-processing in a few cases.
    However, we believe that for harder detection tasks, the 
    benefits of LACBoost would be more impressive.

    The error reduction results of FisherBoost and LACBoost in 
    Fig.~\ref{fig:ROC1} are not as great as those in Fig. \ref{fig:node1}.
    This might be explained by the fact that the
    cascade and negative data bootstrapping remove
    of the error reducing effects, to some extend.
        We have also compared our methods with the boosted greedy sparse LDA (BGSLDA) in
    \cite{Paisitkriangkrai2009CVPR}, which is considered one of the state-of-the-art.
   %
   %
    We provide the ROC curves in the supplementary package.
    Both of our methods outperform 
    BGSLDA with AdaBoost/AsymBoost by about $ 2\%$ in the detection rate. 
    Note that BGSLDA uses the standard cascade.  
    So besides the benefits of our FisherBoost/LACBoost,
    the multi-exit cascade also brings effects.

%
%
%
%
%

\comment{
\begin{figure}[t]
    \begin{center}
        \includegraphics[width=.45\textwidth]{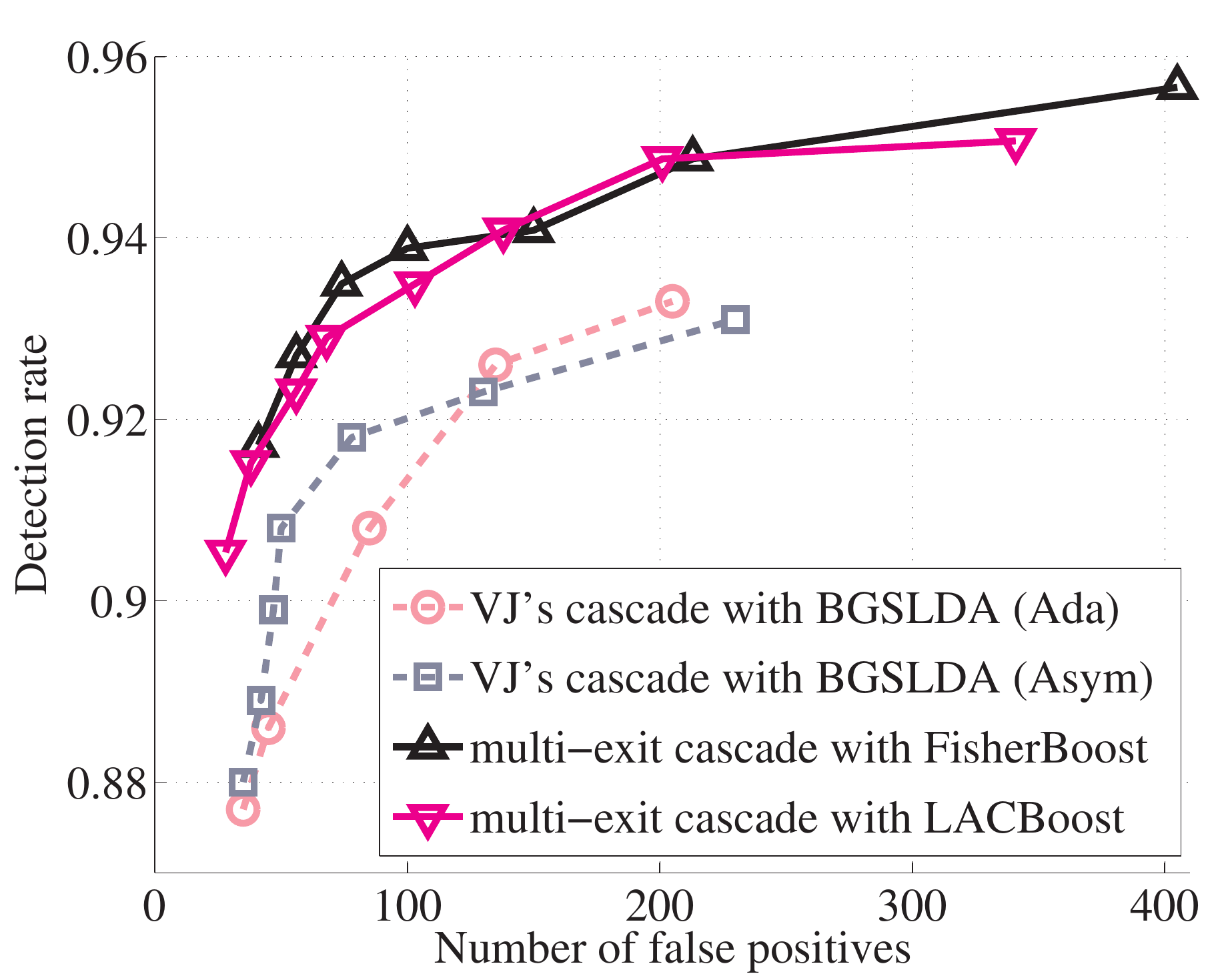}
    \end{center}
    \caption{Cascade performances on the MIT+CMU test data. We compare our methods with BGSLDA in
    \cite{Paisitkriangkrai2009CVPR}.}
    \label{fig:ROC_BGSLDA}
\end{figure}
}

\section{Conclusion}

    By explicitly taking into account the node learning goal in cascade classifiers,
    we have designed 
    new boosting algorithms for more effective object detection. 
    Experiments validate the superiority of our FisherBoost and LACBoost. 
    We have also proposed to use entropic gradient to efficiently 
    implement FisherBoost and LACBoost. The proposed algorithms are easy to implement
    and can be applied other asymmetric classification tasks in computer vision.
    We are also trying to design new asymmetric boosting algorithms
    by looking at those asymmetric kernel classification methods.

\bibliographystyle{splncs}

\end{document}